\documentclass[letterpaper, 10 pt, conference]{ieeeconf}
\IEEEoverridecommandlockouts
\let\labelindent\relax
\usepackage{graphicx} \graphicspath{ {figures/} }
\usepackage{amsmath,amssymb,mathabx}
\usepackage{algorithmic}
\usepackage[ruled]{algorithm2e}
\usepackage{acronym}
\usepackage{enumitem}
\usepackage{booktabs}
\usepackage{hyperref}
\usepackage{balance}
\usepackage{xspace,setspace}
\usepackage[skip=3pt,font=small]{subcaption}
\usepackage[skip=3pt,font=small]{caption}
\usepackage[dvipsnames]{xcolor}
\usepackage[capitalise]{cleveref}
\usepackage{tabularx,colortbl,multirow,array,makecell}
\usepackage{overpic}
\usepackage{cite}
\usepackage{tikz}
\usepackage{fixltx2e}
\usepackage{xcolor}

\makeatletter
\DeclareRobustCommand\onedot{\futurelet\@let@token\@onedot}
\def\@onedot{\ifx\@let@token.\else.\null\fi\xspace}
\def\eg{\emph{e.g}\onedot} 
\def\ie{\emph{i.e}\onedot}

\def\wrt{w.r.t\onedot} 

\makeatother

\crefname{algocf}{alg.}{algs.}
\Crefname{algocf}{Algorithm}{Algorithms}

\makeatletter
\def\BState{\State\hskip-\ALG@thistlm}
\makeatother

\makeatletter
\renewcommand{\paragraph}{%
  \@startsection{paragraph}{4}%
  {\z@}{0ex \@plus 0ex \@minus 0ex}{-1em}%
  {\hskip\parindent\normalfont\normalsize\bfseries}%
}
\makeatother

\crefname{algocf}{alg.}{algs.}
\Crefname{algocf}{Algorithm}{Algorithms}

\definecolor{gblue}{HTML}{4285F4}
\definecolor{gred}{HTML}{DB4437}

\acrodef{dof}[DoF]{Degree of Freedom}
\acrodef{vkc}[VKC]{Virtual Kinematic Chain}
\acrodef{tamp}[TAMP]{Task and Motion Planning}
\acrodef{pddl}[PDDL]{Planning Domain Definition Language}
\acrodef{rrt}[RRT]{Rapidly-exploring Random Tree}
\acrodef{ompl}[OMPL]{Open Motion Planning Library}
\acrodef{iws}[IWS]{Iterated Width Search}
\acrodef{bfs}[BFS]{Breadth First Search}
\acrodef{ai}[AI]{Artificial Intelligence}
\acrodef{spt}[SPT]{Scene Parse Tree}
\acrodef{com}[CoM]{Center of Mass}
\acrodef{mcmc}[MCMC]{Markov Chain Monte Carlo}
\acrodef{ged}[GED]{Graph Editing Distance}
\acrodef{cg}[$cg$]{\textit{Contact Graph}}
\acrodef{cg+}[$cg^+$]{\textit{Contact Graph}$^+$}

\title{\LARGE \bf Sequential Manipulation Planning on Scene Graph\vspace{-9pt}}

\author{Ziyuan Jiao$^{1,2}$\quad{}Yida Niu$^{2}$\quad{}Zeyu Zhang$^{1,2}$\quad{}Song-Chun Zhu$^{1,2,3,4,5}$\quad{}Yixin Zhu$^{2,3,4}$\quad{}Hangxin Liu$^{2\dagger}$% <-this % stops a space
\thanks{$^{1}$ UCLA Center for Vision, Cognition, Learning, and Autonomy (VCLA). 
}%
\thanks{$^{2}$ Beijing Institute for General Artificial Intelligence (BIGAI).
}%
\thanks{$^{3}$ Institute for Artificial Intelligence, Peking University.
}%
\thanks{$^{4}$ School of Artificial Intelligence, Peking University.}
\thanks{$^{5}$ Department of Automation, Tsinghua University.}%
\thanks{$\dagger$ Corresponding author.}
\thanks{Emails: \tt{zyjiao@ucla.edu, niuyida@bigai.ai, zeyuzhang@ucla.edu, sczhu@stat.ucla.edu, yixin.zhu@pku.edu.cn, liuhx@bigai.ai}}
}

\begin{document}

\maketitle
\thispagestyle{empty}
\pagestyle{empty}

\begin{abstract}
We devise a 3D scene graph representation, \emph{contact graph}$^+$ ($\pmb{cg}^+$), for efficient sequential manipulation planning. Augmented with predicate-like attributes, this contact graph-based representation abstracts scene layouts with succinct geometric information and valid robot-scene interactions. Goal configurations, naturally specified on contact graphs, can be produced by a genetic algorithm with a stochastic optimization method. A task plan is then initialized by computing the \ac{ged} between the initial contact graph and the goal configuration, which generates graph edit operations corresponding to possible robot actions. We finalize the task plan by imposing constraints to regulate the temporal feasibility of graph edit operations, ensuring valid task and motion correspondences. In a series of simulated and real experiments, robots successfully complete complex sequential object rearrangement tasks that are difficult to specify using conventional planning language like \ac{pddl}, demonstrating high potential of planning sequential manipulation tasks on $\pmb{cg}^+$.  
\end{abstract}

\setstretch{0.94}

\section{Introduction}

Autonomous robots, expected to conduct a wide range of complex sequential manipulation tasks in challenging environments, ought to have adept planning capabilities. At the task level, robots need to search for a feasible action sequence in a domain, critical for long-horizon tasks involving multiple steps. At the motion level, robots have to produce continuous trajectories by incorporating physical constraints. Yet to date, thoughtfully defining the planning domain at the task level while clearly specifying environmental states at the motion level remains a time-consuming and error-prone process with conventional methods. Despite excelling in expressing symbolic states and abstract actions, STRIPS-like representations (\eg \ac{pddl}) struggle with continuous states like geometric information obtained by the perception module. This deficiency calls for alternative approaches other than STRIPS-like planners, especially for long-horizon manipulation tasks involving complex, nested specifications.

Recently, 3D scene graph emerges as a holistic scene representation for scene modeling~\cite{zhu2007stochastic,zhao2011image,zhao2013scene,qi2018human,jiang2018configurable,huang2018holistic,chen2019holistic++,armeni20193d,qi2020generalized,jia2020lemma,han2021reconstructing,wu2021scenegraphfusion,rosinol2021kimera}, object part modeling~\cite{chang2015shapenet,weng2021captra}, kinematic relations~\cite{huang2021multibodysync,jain2021screwnet}, robot manipulations~\cite{zhu2015understanding,zhu2016inferring,edmonds2017feeling,liu2019mirroring,zhang2020graph}, and human-robot teaming~\cite{liu2018interactive,edmonds2019tale,yuan2020joint}. In particular, \ac{cg}~\cite{han2021reconstructing} reflects the whole kinematic relations detected in the scene using 3D vision, useful for robot \textit{motion} planning~\cite{jiao2021virtual}. In this paper, we further identify that such a \ac{cg} can also serve as a description of \textit{tasks}, thus becoming a carrier of various information related to both task domains and motion constraints. Since \ac{cg} can be directly and robustly built from perceptual input~\cite{han2021reconstructing}, planning manipulation tasks on \ac{cg} naturally bridges robot perception and execution by organizing scene entities, effectively anticipates action outcomes by updating graph, and easily validates physical feasibility by maintaining geometric information.

\begin{figure}[t!]
    \centering
    \begin{subfigure}[b]{\linewidth}
        \centering
        \includegraphics[width=\linewidth]{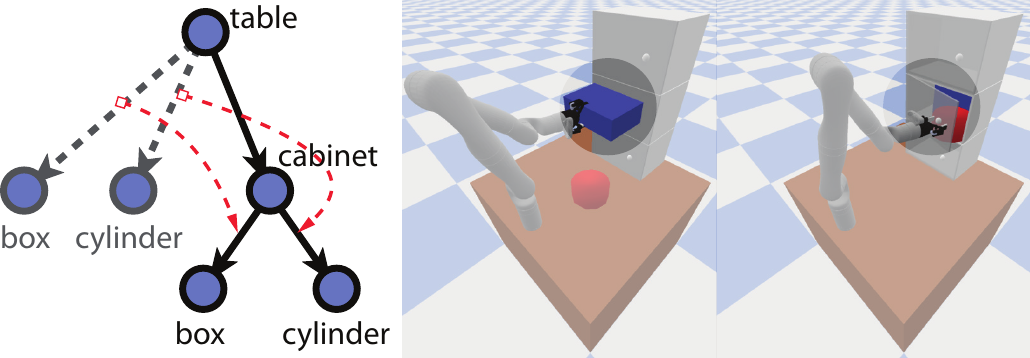}%
        \caption{Problems: Opening the cabinet door must precede placing an object into the cabinet, and two objects must be placed in certain ways such that they do not interfere subsequent operation, \ie, closing the door.}
        \label{fig:moti_init}
    \end{subfigure}%
    \\%
    \begin{subfigure}[b]{\linewidth}
        \centering 
        \includegraphics[width=\linewidth]{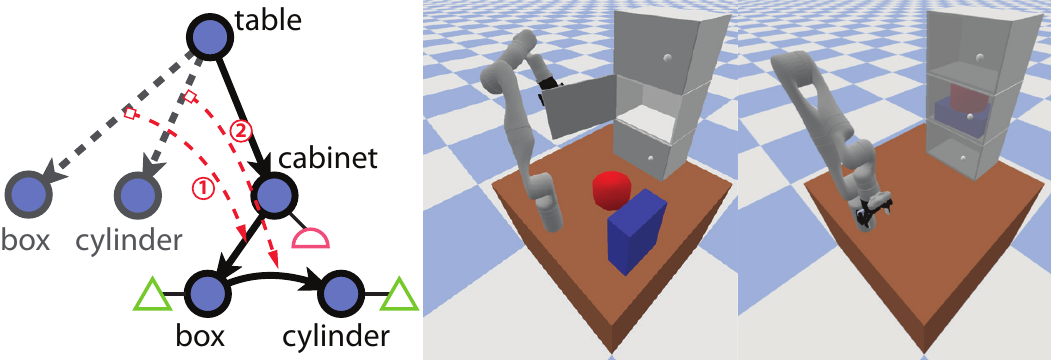}%
        \caption{Solutions: We introduce predicate-like node attributes on \ac{cg}, called \ac{cg+}, to ensure task and motion feasibility.}
        \label{fig:scene_init}
    \end{subfigure}%
    \caption{\textbf{An exemplar problem with the proposed solution using \ac{cg+}.} (a) Planning a complex sequential manipulation task is challenging for conventional task planner. Using contact graph~\cite{han2021reconstructing}, a naive method to plan on graph (\eg, simply using \acf{ged}) faces two challenges\textemdash{}finding the temporal sequence of edit operations and the exact goal configuration. (b) By attaching extra information on nodes as attributes to constrain the problem, the \ac{cg+}-based new framework can generate a valid plan.}
    \label{fig:motivation}
\end{figure}

Given the current environment represented as a \ac{cg}, if one could also specify the goal configuration on a \ac{cg}, a straightforward idea to generate a task plan is to adopt \acf{ged}. Specifically, a \ac{ged} algorithm finds a set of graph edit operations (\eg, inserting and deleting edges) to transform a graph into another. A sequence of feasible graph edit operations naturally corresponds to a set of robot actions, forming a task plan. For instance, deleting an edge and inserting a new one is analogical to picking an object and placing it elsewhere; see \cref{fig:motivation}. However, two challenges have to be addressed.

(i) \textbf{Temporal Dependency.} Some graph edit operations (or robot actions) are invalid until certain prerequisites are met. For instance, in \cref{fig:motivation}a, to represent the task of putting the Box in the Cabinet, it is valid to delete the edge between \texttt{Table} and \texttt{Box} and insert one below \texttt{Cabinet} in graph editing. However, such an operation is infeasible neither in task nor in motion before the cabinet door is open.

\setstretch{0.97}

(ii) \textbf{Goal Configuration.} Computing \ac{ged} requires a valid graph representing the goal configuration. How do we validate whether the goal is physically plausible and produce alternatives when it is not? \cref{fig:motivation}a depicts a scenario where the \texttt{Cabinet}'s volume cannot fit the \texttt{Box} when the \texttt{Cylinder} is placed side by side; valid solutions only exist if one is placed on top of the another.

To tackle these two challenges, we first extend the \ac{cg} to \ac{cg+} (see \cref{fig:scene_init}) by augmenting predicate-like attributes to constrain feasible operations. These attributes are task-specific: They could be predicates/rules in conventional task planners or entities' geometric descriptions naturally defined on graphs. Next, we devise a genetic algorithm for graph structure and a stochastic optimization method for object poses to construct the goal configuration.
To solve the temporal dependency problem, we develop a topological sorting algorithm based on \ac{ged} to search for a sequence of graph edit operations on \ac{cg+} constrained by nodes' attributes, corresponding to the robot's task plan.

In simulation, we demonstrate the proposed graph-based planning scheme in complex sequential manipulation tasks. An experiment further verify the feasibility of the produced task and motion plans in physical environments.
Our contributions are three-fold: (i) Our augmented graph-based representation \ac{cg+} abstracts symbolic forms from 3D perceptual input for task planning while maintaining geometric information for motion planning. (ii) We devise a suite of efficient algorithms for planning complex sequential manipulation tasks on \ac{cg+}. (iii) We demonstrate the potential of using scene graphs as a general representation to organize multiple information sources (\eg, perception, expert knowledge, predicates) and to unify scenes, tasks, and goals.

\subsection{Related Work}

Many effective representations or programming languages have been devised for \textbf{Task Planning}, such as STRIPS~\cite{fikes1971strips}, hierarchical task network~\cite{nau2003shop2}, temporal and-or-graph~\cite{qi2020generalized,edmonds2019tale,liu2019mirroring,qi2018generalized}, Markov decision process~\cite{bellman1957markovian}, and \acs{pddl}~\cite{mcdermott1998pddl}. Among them, \acs{pddl} is a milestone that standardizes task planning. However, \acs{pddl} requires thoughtful designs for complex tasks, which in some cases could become complicated in large planning domains. Although newer versions of \acs{pddl}~\cite{fox2003pddl2,edelkamp2004pddl2,gerevini2005plan} introduced new features to consider more complex planning domains and problems or simplify the domain specification, it is primarily restricted to discrete symbolic variables. Although one of the most up-to-date \acp{pddl}, PDDLStream~\cite{garrett2020pddlstream}, incorporates sampling scheme to deal with high-dimensional and continuous variables, it still requires sophisticated domain-specific functions to process geometric information for related predicates during planning.

While task planning~\cite{karpas2020automated} or motion planning~\cite{lavalle2006planning} alone could be effectively solved nowadays, integrating these two into \textbf{\ac{tamp}}~\cite{garrett2020integrated} remains challenging. Researchers attempt to tackle this problem from various angles, such as incorporating motion-level constraints to the task planning~\cite{erdem2011combining,kaelbling2011hierarchical,garrett2018ffrob,garrett2020pddlstream}, developing interfaces that communicate between task and motion~\cite{srivastava2014combined}, or inducing abstracted modes from motions~\cite{toussaint2015logic,toussaint2018differentiable}. One of the most critical questions is how to scale up for more complex tasks or environments. Our work demonstrates the feasibility of task planning on scene graph representations that naturally represent environmental states, objects geometry, and task goals, efficient for instantiating task plans to motion level.

\subsection{Overview}

We organize the remainder of this paper as follows. \cref{sec:representation} introduces the proposed graph-based representation, \ac{cg+}, and defines attributes. \cref{sec:goal} details the method for goal configuration synthesis, and \cref{sec:planning} introduces the proposed planning schema built on top of \ac{cg+} with \ac{ged}. \cref{sec:exp} further verifies the efficacy of \ac{cg+} scaling to high-dimensional and complex environment. We conclude the paper with discussion in \cref{sec:conclusion}.

\section{Graph-based Scene Representation}\label{sec:representation}

Building on top of \ac{cg}~\cite{han2021reconstructing} representing a 3D indoor scene, \ac{cg+} is augmented as attributes with extra contextual cues for robot planning on complex sequential manipulation tasks.

\begin{figure}[t!]
    \centering
    \includegraphics[width=\linewidth]{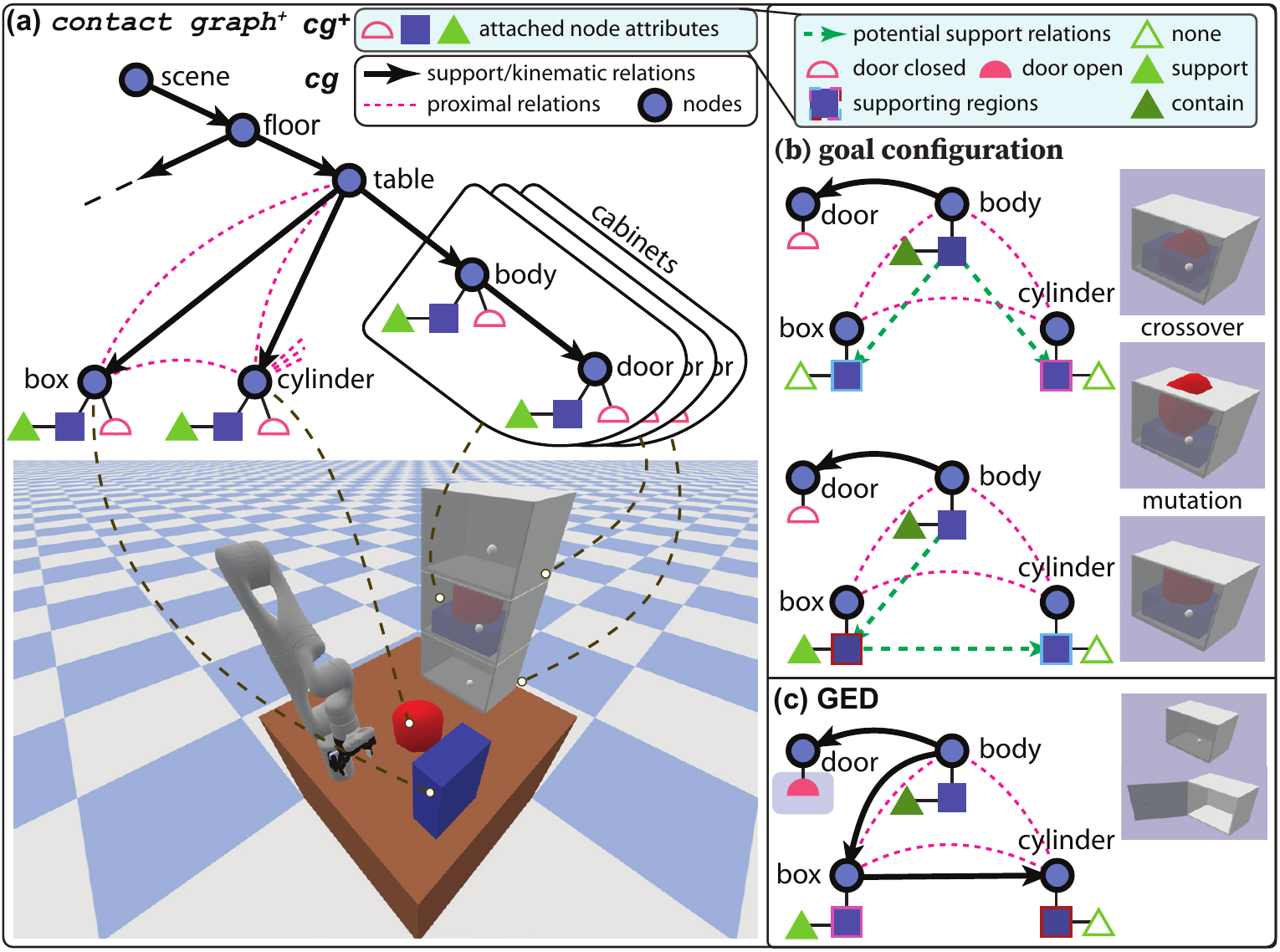}
    \caption{\textbf{Scene representation.} (a) An augmented contact graph \ac{cg+} incorporates additional contextual information as node attributes for robot task planning and motion planning. (b) \textbf{Goal Configuration}: Based on objects' supporting attributes, \ac{cg+} represents a goal configuration that fits two objects into a cabinet. A genetic algorithm is adopted to transform an invalid goal configuration (upper) into a valid one (lower). (c) \textbf{Temporal Dependency}: After generating a set of graph edit operations using \ac{ged}, we further impose a set of object's status attributes as constraints, such that a temporally corrected task plan is generated. Of note, in this scenario, none of the cabinet's descendants in \ac{cg+} are editable when the door's status attribute suggesting that the cabinet door is closed.}
    \label{fig:representation}
\end{figure}

\subsection{Contact Graph}

Formally, a $cg = (pt, E)$ includes (i) a scene parse tree $pt=(V, S)$ that hierarchically organizes scene entities $v_i \in V$ (\eg, objects with their articulated parts) based on supporting relations $S$, and (ii) proximal relations (\eg, collision) $E$ among entities represented by undirected edges. 

\textbf{Scene Entity Nodes}
$V$ include: (i) the scene node $v_s$, serving as the root of $pt$, and (ii) a set of non-root nodes $v_i = \langle o_i, c_i, M_i, B_i \rangle$; each encodes a unique instance label $o_i$, a semantic label $c_i$, a full geometry set of geometry primitives $M_i=\{m_i^j, j=1,\ldots,|M_i|\}$ (a triangular mesh or a CAD model), and an oriented 3D bounding box $B_i$.

\setstretch{0.96}

\textbf{Supporting Relations}
$s_{p,c} \in S$ is a directed edge between the parent node $v_p$ and the child node $v_c$: $s_{p,c}=\langle v_p, v_c \rangle$, indicating $v_p$ stably supports $v_c$ with sufficient contact areas.

\textbf{Proximal Relations}
$E$ introduce links among entities in the $pt$. It imposes additional constraints by modeling spatial relations between two non-supporting but physically nearby objects: Their meshes should not penetrate each other. Proximal relations are only assigned to geometry pairs to enable collision checking and reduce computational costs. The non-penetration constraints are triggered when finding geometrically feasible object poses: 
\begin{equation}
    \text{sd}(m_i, m_j) > 0, \forall (m_i, m_j) \in \mathcal{M}, i\neq j
    \label{eqn:geom_feasible}
\end{equation}
where $\text{sd}(m_i, m_j)$ is the signed distance between $m_i$ and $m_j$~\cite{schulman2014motion}, and $\mathcal{M}$ is a set of all geometry primitive pairs $(m_i, m_j)$ for collision detection. 

\subsection{\texorpdfstring{\acf{cg+}}{}}\label{sec:cg+}

Representing 3D environments by \ac{cg}~\cite{han2021reconstructing} is insufficient to support planning due to the lack of temporal dependency and goal configuration. Here, we augment it to $cg^+ = (pt, E, A)$; see \cref{fig:representation}. While $pt$ and $E$ follow the aforementioned definitions, $A=\{A_i, i \leq |V|\}$ is the set of task-dependent attributes with $A_i$ augmented to a scene entity node $v_i\in V$, which constrain the possible interactions with the node. 

Henceforth, we consider an object rearrangement task with two attributes: (i) a \textbf{supporting attribute} $a^s$ indicates how objects physically support others, and (ii) a \textbf{status attribute} $a^c$ indicates a container's accessibility. Similar to predicates in \acp{pddl}, the attributes of nodes in \ac{cg+} can carry more sophisticated information for other complex tasks.

\begin{figure}[t!]
    \centering
    \includegraphics[width=\linewidth]{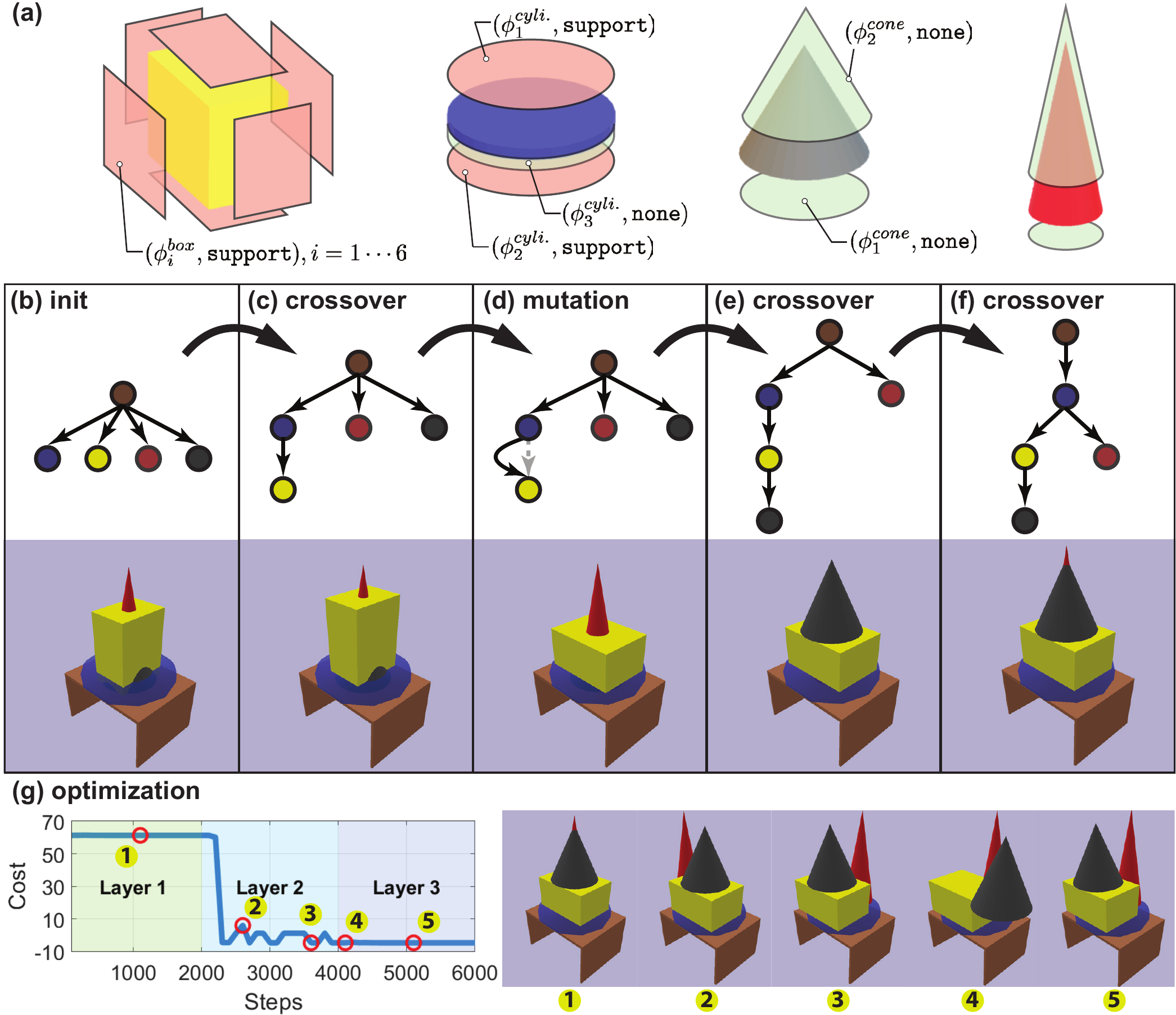}
    \caption{\textbf{Synthesize plausible configuration of placing four objects on the table.} (a) Object's supporting attributes: The box supports others by any of its surface, the disk by its top and bottom, but the cone cannot support others. (b)--(f) The genetic algorithm starts from a rough configuration and searches for a plausible supporting relation by crossover and mutation operations. (g) Objects' poses are further synthesized after the structure is found.}
    \label{fig:support}
\end{figure}

\cref{fig:support}a shows examples of \textbf{supporting attributes} for different shapes. Specifically, the supporting attribute $a^s\subseteq\Phi$ indicates if an object $v$ can support another, where $\Phi=\{(\phi_n, t_n), n = 1,\ldots,|\Phi|\}$ is a set of surfaces extracted from $v$ that would possible serve as a supporting surface. $\phi_n(\cdot)$ is modeled as a region $\Omega_n$ with a closed boundary $\partial\Omega_n$ and is realized as a 2D signed distance field $\phi_n:\mathbb{R}^2\rightarrow\mathbb{R}$, where $\phi_n(\mathbf{x})<0$ is in the interior region $\Omega_n^-$, $\phi_n(\mathbf{x})>0$ in the exterior region $\Omega_n^+$, and $\phi_n(\mathbf{x})=0$ on the boundary $\partial\Omega_n$~\cite{osher2004level}. The value of $\phi_n(\mathbf{x})$ is the minimum Euclidean distance from $\mathbf{x}$ to $\partial\Omega_n$. $t_n\in\{\texttt{none}, \texttt{support}, \texttt{contain}\}$ are the supporting types:
\begin{itemize}[leftmargin=*,noitemsep,nolistsep,topsep=0pt]
    \item \texttt{support} indicates that a stable supporting relation can be formed between $v_p$ and $v_c$ if two nodes satisfy
\begin{equation}
    \phi_{p \cap c}(\text{proj}^\perp_{g}\mathbf{p}_c^\text{com})<0 \iff Stable(s_{p,c}),
    \label{eqn:stab_supp}
\end{equation}
    where $\phi_{p\cap c}$ is defined as a convex hull of an overlapping region between a child node $v_c$ and its parent node $v_p$'s supporting region $\phi_p$. $\text{proj}^\perp_{g}\mathbf{p}_c^\text{com}$ projects \ac{com} of $v_c$ and all its descendants onto the same plane as $\phi_{p\cap c}$, and $g$ is the gravitational vector; we assume all supporting planes are perpendicular to $g$. $\phi_{p \cap c}>0$ indicates the gravitational moment exerted on $v_c$ cannot be canceled by its support and result in unstable $s_{p,c}$.
    \item \texttt{contain} is a step further to \texttt{support}. In addition to satisfying \cref{eqn:stab_supp}, the bounding box of $v_p$ should contain the union of bounding boxes for $v_c$ and all its descendants:
\begin{equation}
    \small
    \text{Vol}\left(\left(\bigcup_{v_i\in st(v_c)} B_i\right) \bigcup B_p\right) = \text{Vol}(B_p),
    \label{eqn:contain}
\end{equation}
    where $st(v_c)$ is a subtree of $pt$ rooted at $v_c$.
    \item \texttt{none} indicates that no support or contain relations could be established between the two nodes. 
\end{itemize}

The \textbf{status attribute} $a^c\in\{\texttt{closed}, \texttt{opened}\}$ determines whether the objects descended from a node with \texttt{contain} attribute (\ie, being contained) are accessible.

$A_i=(a_i^s, a_i^c)$ is the assigned attribute to node $v_i$, wherein $a_i^c$ is an optional attribute only assigned to containers. The supporting relation $s_{p,c}$ is further augmented with supporting region $\phi_p$ and kinematic information $\mathbf{p}_{p,c}$: $s_{p,c}=\langle v_p, v_c, \mathbf{p}_{p,c}, (\phi_p, t_p) \rangle$. $\mathbf{p}_{p,c}\in\mathbb{R}^{n_{p,c}}$ is the pose vector pointing from $v_p$ to $v_c$, where $0\leq n_{p,c}\leq3$ is the \ac{dof} of $v_c$ \wrt $v_p$; it could also be considered as the plausible transformation between the parent and the child nodes. $(\phi_p, t_p)\in a_p^s$ is a supporting attribute from the parent node $v_p$ to establish the supporting relation $s_{p,c}$. Of note, a single node is possible to have more than one supporting attribute in $a_i^s$, which affords to partition working space on the same node for more sophisticated tasks.

\subsection{Problem Definition}

We define the problem of planning on \ac{cg+} in two phases. Assuming a rough goal configuration $cg_g^{+\prime}$ is provided (\eg, putting the box and the cylinder into the cabinet; see \cref{fig:representation}bc), the first phase resolves the violations of physical constraints defined in \cref{sec:cg+}. To modify $cg_g^{+\prime}$ and discover a plausible \ac{cg+}, we integrate a genetic algorithm to produce the structure of supporting relations among objects and a stochastic optimization method to generate their poses that satisfy physical constraints imposed on both $S$ and $E$. \cref{sec:goal} describes goal configuration discovery.

\setstretch{0.99}

We utilize \ac{ged} to find an optimal set of graph edit operations $e_{1:T}=(e_1,\ldots,e_T)$ to transform from the current environment represented by $cg_0^+$ to the goal configuration $cg_g^+$, along with imposed temporal dependencies between operation pairs (\eg, opening cabinet door before placing objects inside); see \cref{fig:representation}d. \cref{sec:planning} details this procedure. 

\section{Goal Configuration Discovery}\label{sec:goal}

A planning process requires a known goal configuration. Although defining one as a \ac{cg+} is relatively straightforward, automatically finding a plausible goal configuration satisfying all physical constraints is still preferred.

Assuming a rough goal is specified (see \cref{fig:support}b), we aim to find a configuration incorporating proper \textbf{supporting} and \textbf{proximal} relations ($S$ and $E$) among objects. For a $cg^+=(pt,E,A)$, its configuration is represented by $S$, and its configurations space $\mathcal{S}$ is defined as $(\mathcal{A}\times\mathcal{A})^{|V|-1}\times\mathbb{R}^N$, where $\mathcal{A}$ is the attribute space, and $N$ is the object poses' total \acp{dof}. Directly sampling a configuration of \ac{cg+} in $\mathcal{S}$ may not always produce valid configurations. We address this problem by using (i) a genetic algorithm to generate a $pt$'s structure in $(\mathcal{A}\times\mathcal{A})^{|V|-1}$ to discover \textbf{supporting} relations, and (ii) a stochastic optimization method to optimize the object poses in $\mathbb{R}^N$ to obtain valid \textbf{proximal} relations.

\subsection{Supporting Structure Synthesis}

Genetic algorithms have demonstrated its capability of searching complex tree structures for symbolic regression~\cite{koza1994genetic,udrescu2020ai}. It consists of two basic operations, \texttt{crossover} and \texttt{mutation}, which randomly modify the edges and nodes over generations to increase diversity in a population. In addition, a fitness function is defined as a heuristic to select preferred tree structure over the population. In this paper, we adopt a genetic algorithm for our $pt$ structure generation. Specifically,
\begin{itemize}[leftmargin=*,noitemsep,nolistsep,topsep=0pt]
    \item \textbf{\texttt{Crossover}} breaks a supporting relation $s_{p,c}$ and transplants $v_c$ with all its descendants to another parent $v_{p'}$, as long as $v_{p'}$ is not $v_c$'s descendant, and the new supporting relation satisfies constraints imposed by the supporting attribute $a_{p'}^s$. \cref{fig:support} illustrates some \texttt{Crossover} operations.
    \item \textbf{\texttt{Mutation}} first randomly selects another set of supporting attributes $a_m^s \subseteq \Phi_m$ for a random node $v_m$. Next, it chooses a $(\phi_m(\cdot), t_m)\in a_m^s$ for possibly better supporting of its descendants or larger space for maintaining proximal relations with its surrounding objects.
\end{itemize}

We design a \textit{Fitness} score $\mathcal{F}$ as a search heuristic to speed up the supporting synthesis:
\begin{align}
     \mathcal{F} = \sum_{s_{p,c}\in S} \text{max}(\frac{Area(\phi_{p\cap c})}{Area(\phi_p)}, \theta) - \theta,
     \label{eqn:fitness}
\end{align}
where $\theta$ is a threshold of area occupation. Intuitively, no more objects can be placed on the parent node if the contact area between the parent node and its child node(s) is larger than $\theta$, and the algorithm would incline to move the child node(s) away. \cref{fig:support}b-f depicts an example of how a valid supporting structure in \ac{cg+} is found by the algorithm. Next, we describe how to find a detailed configuration with specified object poses.

\setstretch{1}

\subsection{Object Pose Synthesis}\label{sec:find_pose}

The objects' poses in a valid \ac{cg+} should satisfy \cref{eqn:geom_feasible}, \ie, not penetrating each other. With a hinge loss function
\begin{align}
    \mathcal{L}^\text{sd}_{i,j}&=\text{max}(0, -\frac{\text{sd}(m_i,m_j)}{d_\text{safe}}+1),
    \label{eqn:loss_sd}
\end{align}
we penalize the signed distance between objects $m_i$ and $m_j$, and $d_{\text{safe}}>0$ is a safety distance among them. $\mathcal{L}^\text{sd}_{i,j}=0$ if $\text{sd}(m_i,m_j)\geq d_{\text{safe}}$, and $\mathcal{L}^\text{sd}_{i,j}\geq1$ if $m_i$ and $m_j$ are in collision. We formulate the object pose synthesis as an optimization problem:
\begin{align}
    \text{minimize}\quad{}
    &\sum_{i}\sum_{j} \mathcal{L}^\text{sd}_{i,j},
    \label{eqn:single_obj}
    \\
    \text{subject to}\quad{}
    &\text{sd}(m_i,m_j) > 0,\quad{} \forall (m_i, m_j) \in \mathcal{M}.
    \label{eqn:single_geom_cnt}
\end{align}
We can further impose constraints for support or contain (\ie, \cref{eqn:stab_supp,eqn:contain}) to this optimization process. To solve this optimization efficiently, we design an update scheme:
\begin{align}
    \pmb{\mu}_c^{\text{sd}}&=\sum_i\sum_j \pmb{\mu}_{i,j}^{\text{sd}}, \ \forall m_i \in M_c,\ (m_i, m_j) \in \mathcal{M} \label{eqn:total_sd_dir}
    \\
    \pmb{\mu}_{i,j}^{\text{sd}}&=\frac{\mathcal{L}^\text{sd}_{i,j}}{\mathcal{L}^\text{sd}_\text{total}}\frac{\textbf{p}_{m_i}-\textbf{p}_{m_j}}{||\textbf{p}_{m_i}-\textbf{p}_{m_j}||_2}.
    \label{eqn:sd_dir}
\end{align}
\cref{eqn:total_sd_dir} is a weighted sum over pose vectors defined in \cref{eqn:sd_dir} between two objects, whose weights are proportional to the signed distance loss (\cref{eqn:loss_sd}). This design implicitly pushes object away to resolve collision or increases safety distance.

We further add stochasticity to avoid local minima. The update direction of optimization is 
    $\mathbf{x}'=\mathbf{x}+\delta\cdot(\mathbf{\pmb{\mu}}+\pmb{\sigma}_\gamma\cdot\mathcal{N}(0,1))$,
where $\delta$ is the step size, $\mathbf{x}$ is a pose vector which is concatenated by objects poses $\mathbf{p}$ for optimization, $\mathbf{\pmb{\mu}}+\pmb{\sigma}_\gamma\cdot\mathcal{N}(0,1)$ is the proposal distribution to be sampled from, and $\mathbf{\pmb{\mu}}$ is the direction for the next sample. $\pmb{\sigma}_\gamma$ adds noise to the sampling direction; it decays at the rate $\gamma\in(0,1)$ in each iteration, which reduces randomness in sampling process as iteration increases. The optimization is realized iteratively in a breadth-first manner; see \cref{alg:structure_syn}. \cref{fig:support}g shows an example of the pose synthesis process. In Layer 1, only one object is in the lowest level (disk) and the highest layer (grey cone), whose poses are found quickly. The convergence is slower in Layer 2 as the box and the red cone should not collide with each other while staying within the disk.

\begin{algorithm}[ht!]
    \small
    \caption{Optimization of object poses over \ac{cg+}}
    \label{alg:structure_syn}
    \LinesNumbered
    \SetKwInOut{KIN}{Input}
    \SetKwInOut{KOUT}{Output}
    \SetKwInOut{Param}{Params}
    \KIN{$pt$ Unoptimized scene parse tree in \ac{cg+}\\
    }
    \KOUT{$pt^*$ Optimized scene parse tree in \ac{cg+}\\
    }
    \For{depth in 0:(MAX\_DEPTH(pt)-1)}{
        $st \leftarrow pt.GetSubtree(pt.root, depth+1)$

        \For{node at depth}{   
            \If{$child(node)\neq \emptyset$}{
                $x \leftarrow st.GetPose(child(node))$
                
                $x^* \leftarrow st.ObjectPoseSynthesis(x)$
            
                $pt.SetTreePose(x^*)$
            }
        }
    }
    $pt^* \leftarrow pt$
\end{algorithm}

\setstretch{0.97}

\section{Planning on \texorpdfstring{\ac{cg+}}{}}\label{sec:planning}

We detail our planning framework based on \ac{cg+} for a single agent (\eg, a single manipulator), assuming a large swap node $v_\text{swap}$ (\eg, a table) is available to temporarily place objects. First, the planning framework leverages \ac{ged} to find an initial action set. Next, the action set and temporal dependencies among actions are constructed incrementally by reasoning about physical commonsense in terms of accessibility, stability, and collision. Finally, a valid action plan is found through the topological sort.

\subsection{Graph Edit Operations}\label{sec:action_definition}

The concept of \ac{ged}~\cite{sanfeliu1983distance} is first introduced to measure the similarity between two graphs. It finds a set of graph edit operations that transform a graph into another while minimizing the total editing cost. We define \ac{ged} between the initial scene $pt_0$ and the goal configuration $pt_g$ as
\begin{equation}
    GED(pt_0, pt_g)=\min_{(e_1,\ldots,e_k) \in P(pt_0, pt_g)}\sum_{i=1}^k c(e_i),
    \label{eqn:ged}
\end{equation}
where $c(e_i)$ is the cost function of an edit operation $e_i$, and $P(pt_0, pt_g)$ is a set of edit operations transforming $pt_0$ to $pt_g$. We consider four types of edit operations and correspond them to robot actions:
\begin{itemize}[leftmargin=*,noitemsep,nolistsep,topsep=0pt]
    \item delete($s_{p,c}$) $\rightarrow$ \texttt{Pick}($v_p$, $v_c$): Pick an object $v_c$ from $v_p$.
    \item insert($s_{p,c}$) $\rightarrow$ \texttt{Place}($v_p$, $v_c$): Place an object $v_c$ on $v_p$.
    \item substitute($a_i^c$, \texttt{opened}) $\rightarrow$ \texttt{Open}($v_i$): Open the door $v_i$ such that edges among contained objects are editable.
    \item substitute($a_i^c$, \texttt{closed}) $\rightarrow$ \texttt{Close}($v_i$): Close the door $v_i$ such that edges among contained objects are uneditable.
\end{itemize}

We use the \ac{ged} algorithm~\cite{abu2015exact} to find $P(pt_0, pt_g)$ containing a set of edit operations that transforms $pt_0$ to $pt_g$; edit operations are referred to as robot actions henceforth.

\subsection{Temporal Dependency}\label{sec:temp_dep}

Although the robot action set $P$ provides elements for planning, generating feasible plans requires valid temporal dependencies. We build up a partially ordered set $(P,\ C)$ by imposing temporal dependencies onto certain pairs of robot actions $C$. We consider three types of dependencies.

\paragraph*{Action Precedence}

Some actions should take place before others. For instance, an object has to be picked before it can be placed: $\texttt{Pick}(\cdot,v) < \texttt{Place}(\cdot,v)$, and the parent object must be placed before placing others on the top of it: $\texttt{Place}(\cdot,v_p) < \texttt{Place}(v_p,v_c)$.    

\paragraph*{Spatial Feasibility}

Some objects should be cleared before performing the action. For instance, re-orientating a parent node (\eg, flip a box upside down) requires all its descendants to be placed elsewhere (\eg, a swap node $v_\text{swap}$):
\begin{equation*}
    \resizebox{\hsize}{!}{$%
        \begin{aligned}
            P\cup\{&\texttt{Pick}(v_c, v_{d_i}), \texttt{Place}(v_\text{swap}, v_{d_i}), \\
            &\texttt{Pick}(v_\text{swap}, v_{d_i}), \texttt{Place}(v_c, v_{d_i}), \forall v_{d_i}\in child(v_c) \}, \\
            \text{where} & \\
            & \texttt{Pick}(v_c, v_{d_i}) < \texttt{Place}(v_p, v_c), \texttt{Place}(v_p, v_c) < \texttt{Place}(v_c, v_{d_i}).%
        \end{aligned}
    $}
\end{equation*}

\paragraph*{Accessibility}

It is prohibited to interact with others inside a closed enclosure, \ie, editing the edges among all its descendants. For instance, an \texttt{Open} action should precede all related graph edit operation in $P$,
$
    \cdots<\texttt{Open}<\texttt{Pick}<\cdots<\texttt{Place}<\texttt{Close}<\cdots.
$

\setstretch{0.935}

\subsection{Topological Sort}

Given $(P,\ C)$, the task planning problem on \ac{cg+} becomes a topological sorting problem on $(P,\ C)$ to produce a valid action sequence $e_{1:T}=(e_0,\ldots,e_T)$. To solve it, we define a search node $\mathcal{N}=(P', C', e', cg^{+'})$, where $P' \subseteq P$ is the set of actions remains unexplored, $C' \subseteq C$ is the set of temporal dependencies not imposed yet, $e'$ the selected action reaching current search node, and $cg^{+'}$ the graph structure after executing $e'$. At the start node, $P$ and $C$ are given by \cref{sec:temp_dep}. $P \setminus C$ contains actions that do not have precursors, available for exploration of neighbors.

The \ac{cg+} representation is advantageous for evaluating geometric feasibility (\eg, collision) during task planning. Specifically, each action is parametrized by the object poses with geometric information encoded in the nodes, enabling collision detection during planning. Compared to \acs{pddl} definitions, which have to checks \textit{all} the pair-wise relations, the \ac{cg+} representation maintains object relations over hierarchical structures and evaluates geometric feasibility \textit{only} on edited node and its related nodes (\eg, descendants). The infeasibility can be resolved by (i) moving the object to the swap node instead of directly to its goal, (ii) adding new actions to $P$ that move the object back to the goal, and (iii) imposing necessary temporal constraints to clear objects that would collide along the way. The unspecified intermediate goals can be found by reiterating the pose synthesis algorithm in \cref{sec:find_pose}. Our searching pipeline is implemented as a depth-first-search-based topological sorting algorithm, and the object poses are optimized along with the collision checking process during the search.

\begin{figure}[b!]
    \centering
    \begin{subfigure}[b]{0.25\linewidth}
        \centering 
        \includegraphics[trim=190 230 190 230,clip,width=\linewidth]{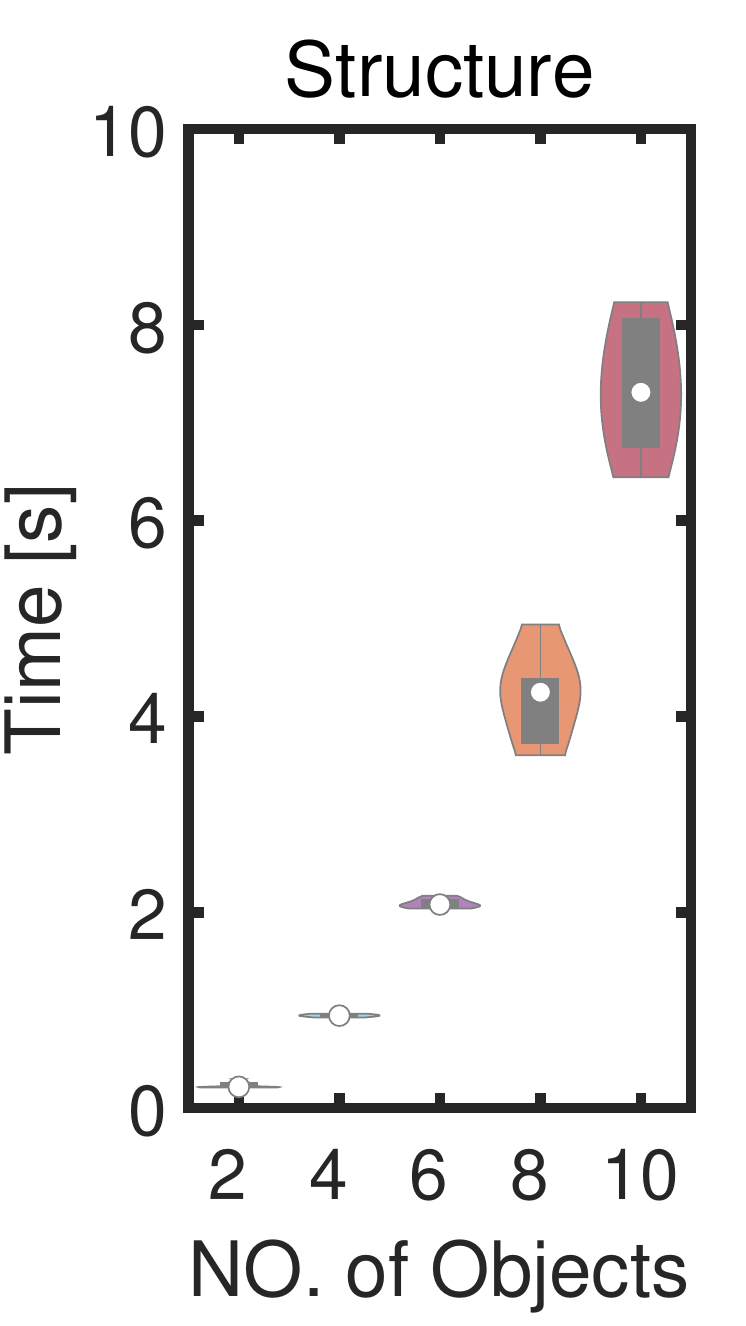}
        \caption{}
        \label{fig:exp1_graph}
    \end{subfigure}%
    \begin{subfigure}[b]{0.25\linewidth}
        \centering 
        \includegraphics[trim=190 230 190 230,clip,width=\linewidth]{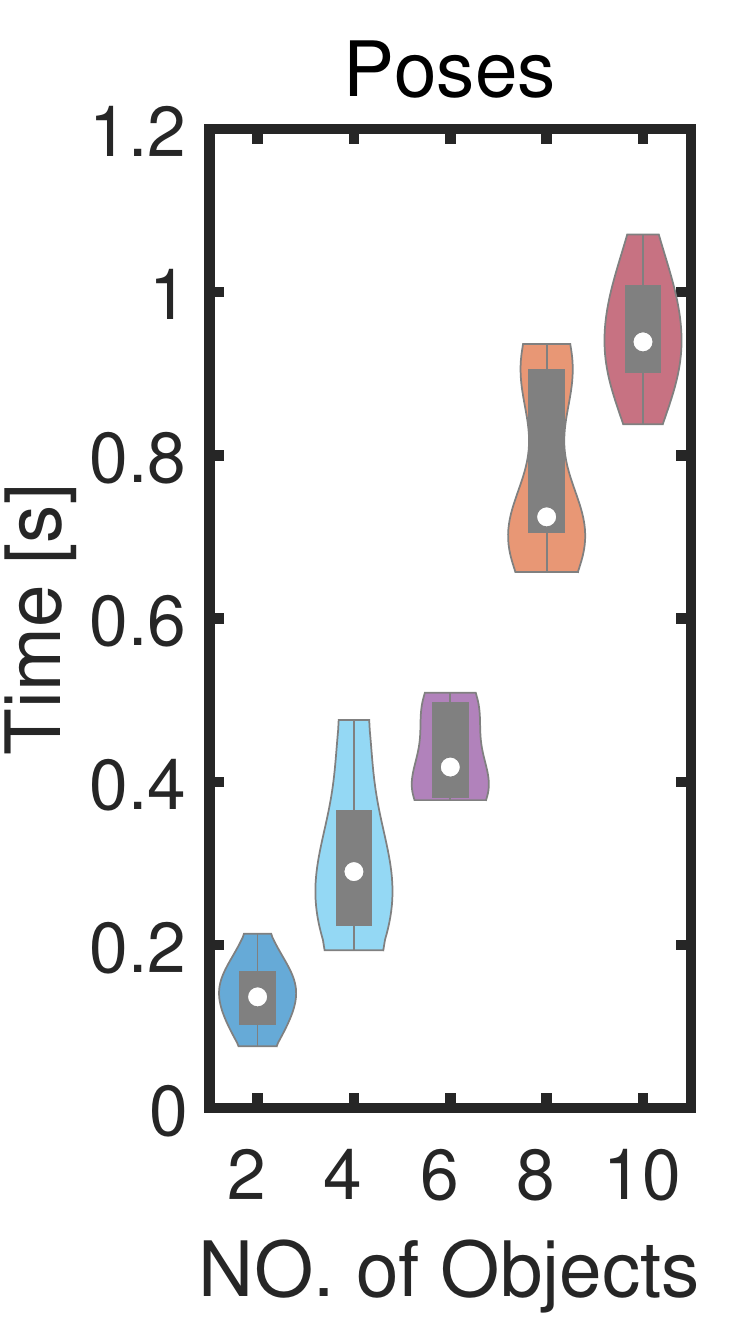}
        \caption{}
        \label{fig:exp1_pose}
    \end{subfigure}%
    \begin{subfigure}[b]{0.25\linewidth}
        \centering 
        \includegraphics[trim=190 230 190 230,clip,width=\linewidth]{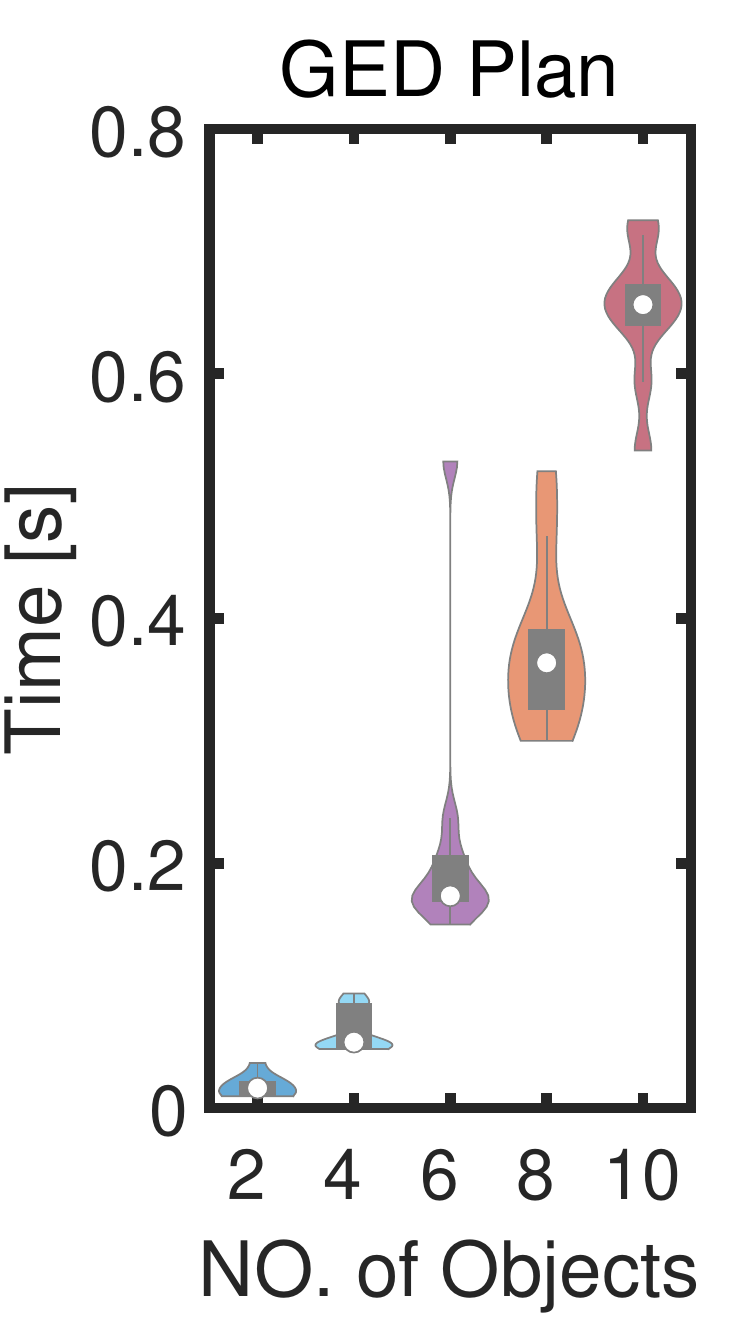}
        \caption{}
        \label{fig:exp1_ged}
    \end{subfigure}%
    \begin{subfigure}[b]{0.25\linewidth}
        \centering 
        \includegraphics[trim=190 230 190 230,clip,width=\linewidth]{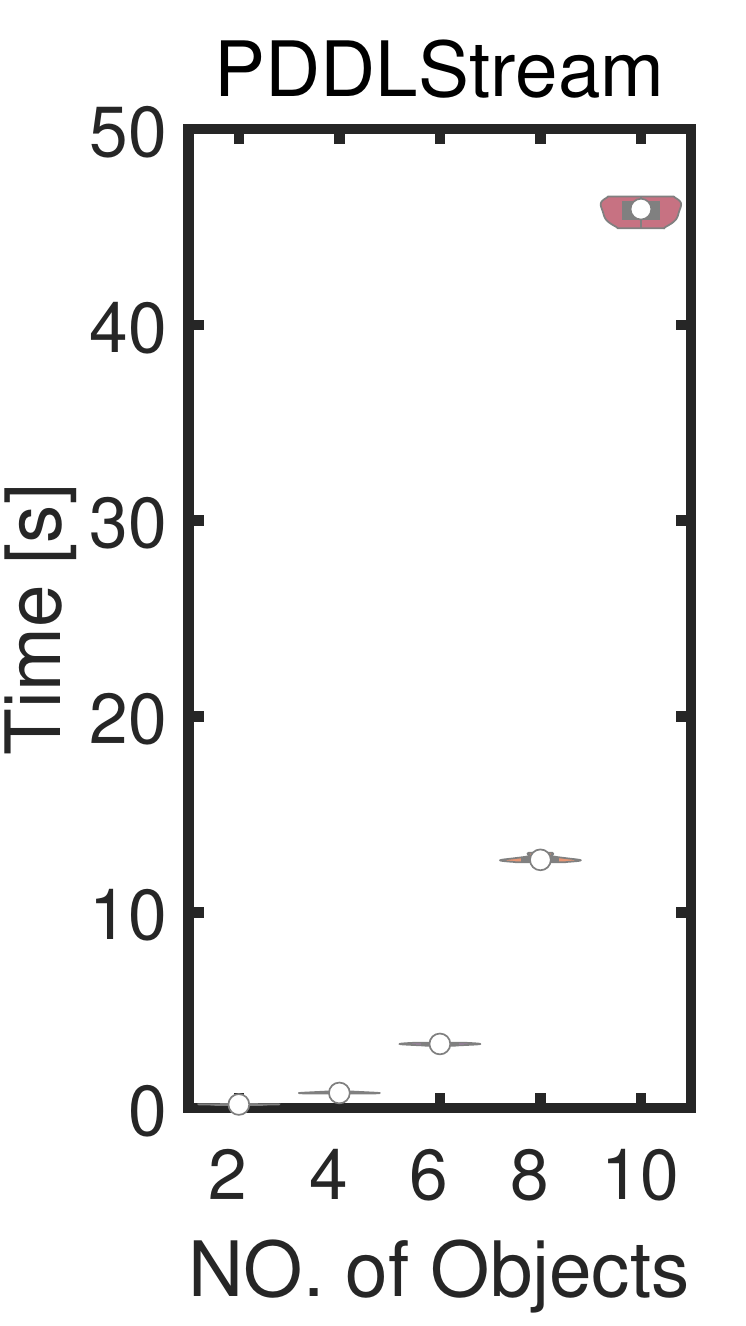}
        \caption{}
        \label{fig:exp1_pddl}
    \end{subfigure}%
    \hfill
    \begin{subfigure}[b]{\linewidth}
        \centering 
        \includegraphics[width=\linewidth]{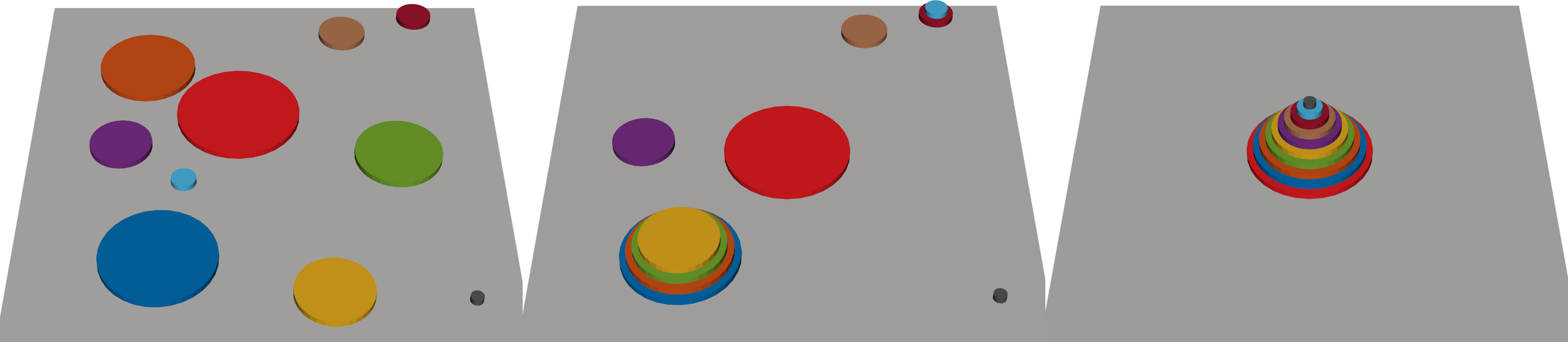}
        \caption{From left to right: Initial, intermediate, and final configurations.}
        \label{fig:exp1_setup}
    \end{subfigure}%
    \caption{\textbf{Stack ten objects.} (a--d) Violin plots~\cite{hintze1998violin} (a hybrid of a box plot and a kernel density plot) of the computing time. The white dot represents the median, the thick gray bar in the center the 25\% to 75\% quartile range, and the color shaded areas the data distribution. (e) Corresponding simulation setup.}
    \label{fig:time}
\end{figure}

\section{Simulations and Experiments}\label{sec:exp}

In simulations, we characterize the algorithms supporting our graph-based planning framework by time complexity in an object stacking task. We further demonstrate that our framework can handle a complex sequential manipulation task. In the experiment, we use a physical robot manipulator in a setup similar to \cref{fig:motivation}. The code and environment are available at \href{https://sites.google.com/view/planning-on-graph}{https://sites.google.com/view/planning-on-graph}.

\setstretch{0.95}

\begin{figure*}[t!]
    \centering
    \begin{subfigure}[b]{\linewidth}
        \centering 
        \includegraphics[width=0.95\linewidth]{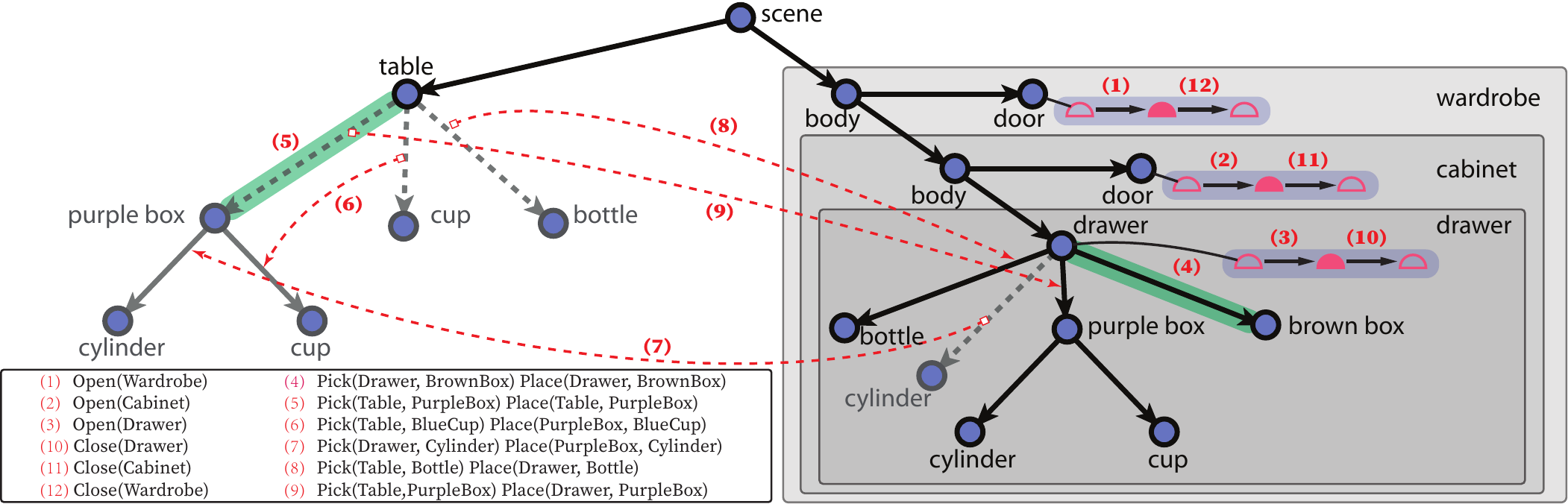}
        \caption{\textbf{Edit operations on $\mathbf{\ac{cg+}}$.} The purple shaded attributes indicate open/closed actions due to the accessibility criteria enforced by temporal dependency. The green shaded edges highlight the \texttt{Mutation} procedure when synthesizing the structure in goal configuration\textemdash{}switching supporting surfaces may better satisfy geometric constraints (\eg, (4)) or is more efficient in subsequent actions (\eg, (5)).}
        \label{fig:exp2_graph}
    \end{subfigure}%
    \hfill
    \begin{subfigure}[b]{\linewidth}
        \centering 
        \includegraphics[width=\linewidth]{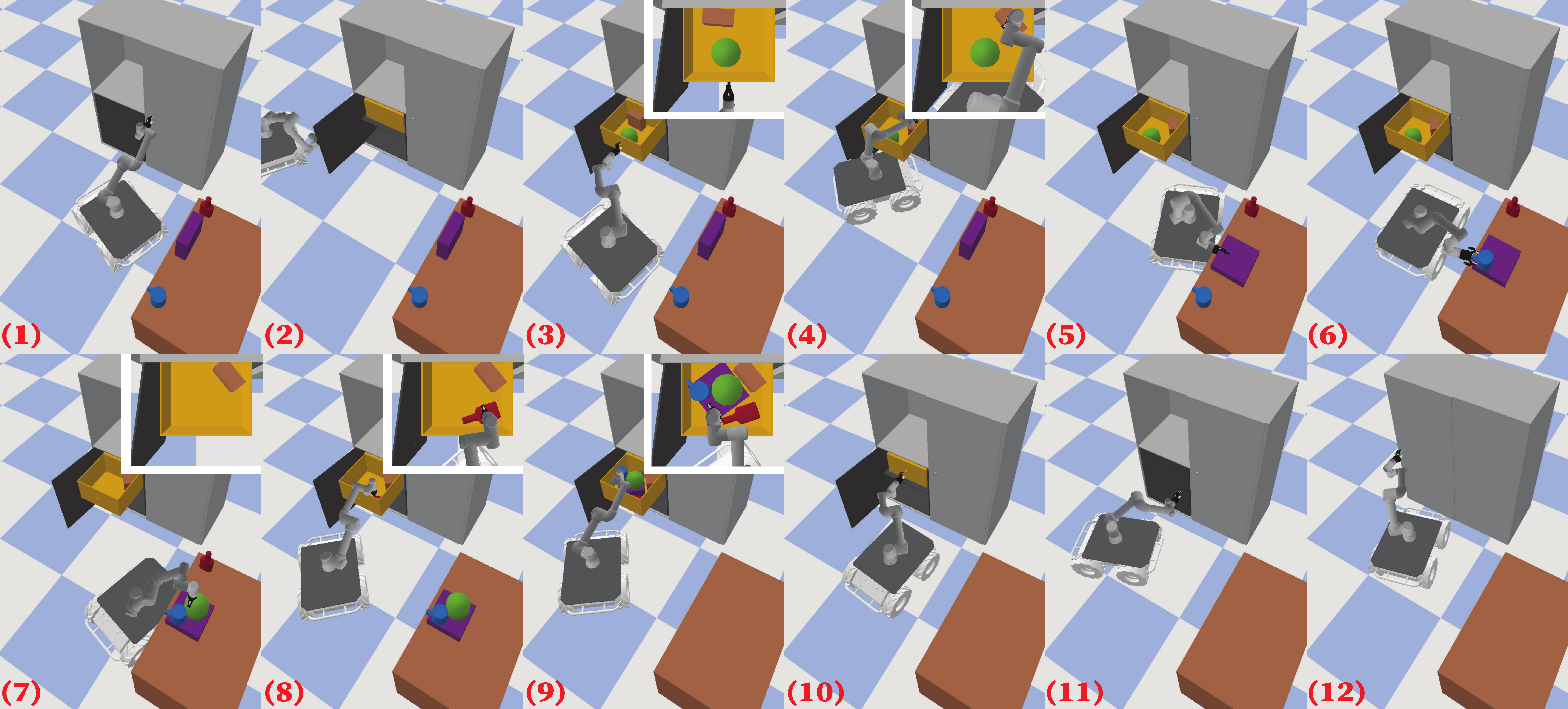}
        \caption{\textbf{Key robot actions performed during the task execution.}}
        \label{fig:exp2_task}
    \end{subfigure}%
    \caption{\textbf{Planning results for a long-horizon and geometrically-complex object rearrangement task.} (a) The graph edit operations and the corresponding robot actions planned by our framework successfully move everything on the \texttt{Table} to the \texttt{Drawer}. (b) Constructed from \ac{cg+}, motion planning on \acs{vkc} plans action sequence, which consolidates the mobile manipulator and the manipulated objects.}
    \label{fig:exp2}
\end{figure*}

\subsection{Simulated Object Stacking}

We design an object-stacking task to evaluate the time complexity of the proposed framework, especially when facing complex scenarios. In this task, an agent stacks, through symbolic actions, the plates on the table from large to small; \cref{fig:exp1_setup} shows a typical example of this task's initial and goal configurations with ten plates.

While the task appears to be simple, stacking each object requires \texttt{Pick} and \texttt{Place} that account for pair-wise constraints in terms of object size. Hence, its complexity quickly increases as the number of objects increases. Our framework needs three steps to solve this problem: (i) synthesizing structure (\ie, supporting relations) among objects from large to small, (ii) synthesizing specific object poses such that the smaller object on top is within the larger one, and (iii) producing a complete task plan by \ac{ged}.

We also implement a baseline using an existing task planning method. Classic \acp{pddl} do not contain object geometric information and fail to handle the pair-wise constraints in terms of object size. Here, we implement a PDDLStream, capable of considering object poses by incorporating samplers in \acs{pddl} domains. We define two streams that (i) sample poses of an object supported by another, and (ii) test collision between two objects given their poses. We choose the default adaptive algorithm provided by the PDDLStream. The goal for both the baseline and our method is implicitly defined as pair-wise constraints in terms of object sizes: A smaller object must be placed above a larger object, resulting in only one valid goal configuration. The simulation is repeated 10 times for 2, 4, 6, 8, and 10 objects with randomly initialized object poses. The computing hardware is a Ubuntu 20.04 desktop with an AMD 5950x processor.

\cref{fig:exp1_graph,fig:exp1_pose,fig:exp1_ged,fig:exp1_pddl} show the simulation results. For the baseline method, the time requirement of finding a valid solution using PDDLStream grows exponentially (\cref{fig:exp1_pddl}) as the number of object increases since it needs to explore all valid combinations of actions and objects in the search space. 

\setstretch{1}

In comparison, the proposed framework achieves the goal through three steps: synthesizing structure of supporting relations (\cref{fig:exp1_graph}), synthesizing object poses (\cref{fig:exp1_pose}), and planning with \ac{ged} (\cref{fig:exp1_ged}). By leveraging the \ac{cg+} representation, the genetic algorithm only explores the graph structures that are more likely to produce a feasible solution, which successfully reduces the search space and is especially advantageous when the setup is more complex. Still, the majority of time is spent on synthesizing structure due to the large space of plausible graph structures. Overall, the proposed method is significantly more efficient than PDDLStream when the environments are complex (\ie, more than six objects), whereas PDDLStream is at similar performance (slightly better) when the scenarios are simple (\ie, six or fewer objects). This result indicates that the proposed framework can scale up much better compared with PDDLStream.

\subsection{Simulated Complex Task}

We qualitatively demonstrate that our graph-based planning scheme can handle a complex sequential manipulation task. The task is to put \texttt{Cup}, \texttt{Bottle}, and \texttt{PurpleBox} on a table inside a drawer. The challenges are two-fold: (i) Since the \texttt{BrownBox} and the \texttt{Cylinder} are already inside the confined \texttt{Drawer}, the objects must be rearranged properly (\ie, synthesize a valid goal configuration) to fit in the tight space. (ii) The \texttt{Drawer} is contained by the \texttt{Cabinet}, who is also contained by \texttt{Wardrobe}, demanding a feasible plan with the correct temporal dependency when reaching for the objects inside the drawer.

\cref{fig:exp2_graph} shows the produced task plan and depicts the corresponding graph edit operations on the \ac{cg+}. Of note, most operations consist of two actions\textemdash{}\texttt{Pick} and \texttt{Place}. The corresponding keyframes in the task execution using a mobile manipulator (a Husky mobile base and a UR5 arm) are visualized in \cref{fig:exp2_task}, whose motions are planned using a \acf{vkc} modeling method~\cite{jiao2021efficient,jiao2021virtual}. 

Resulting produced using the \ac{cg+} representation reflect three advantages.
First, the robot produces a temporally correct plan by exposing three levels of containment and closing them afterward; see \cref{fig:exp2_task}(1--3) and (10--12). These nested relations are not trivial to define by predicates; they are naturally expressed in \ac{cg+} due to its hierarchy.
Second, the plan contains critical steps of rearranging objects. For instance, move the \texttt{BrownBox} to the side in the \texttt{Drawer} (\cref{fig:exp2_task}(4)) and place the \texttt{Bottle} flat (\cref{fig:exp2_task}(8)), so that the large \texttt{PurpleBox} could fit in the confined space. These capabilities demonstrate the efficacy of goal configuration synthesis, facilitated by the geometric information encoded in \ac{cg+}. 
Third, the task plan is very efficient\textemdash{}the robot can make the best use of the \texttt{PurpleBox} to support the \texttt{Cup} and the \texttt{Cylinder}, so that they can be moved together and minimize motion costs. Achieving this using PDDLStream would require tedious definitions of object pair-wise relations and complex descriptions of objects for sampling.  

\subsection{Experiment}

\begin{figure}[t!]
    \centering
    \begin{subfigure}[b]{0.3\linewidth}
        \centering 
        \includegraphics[width=\linewidth]{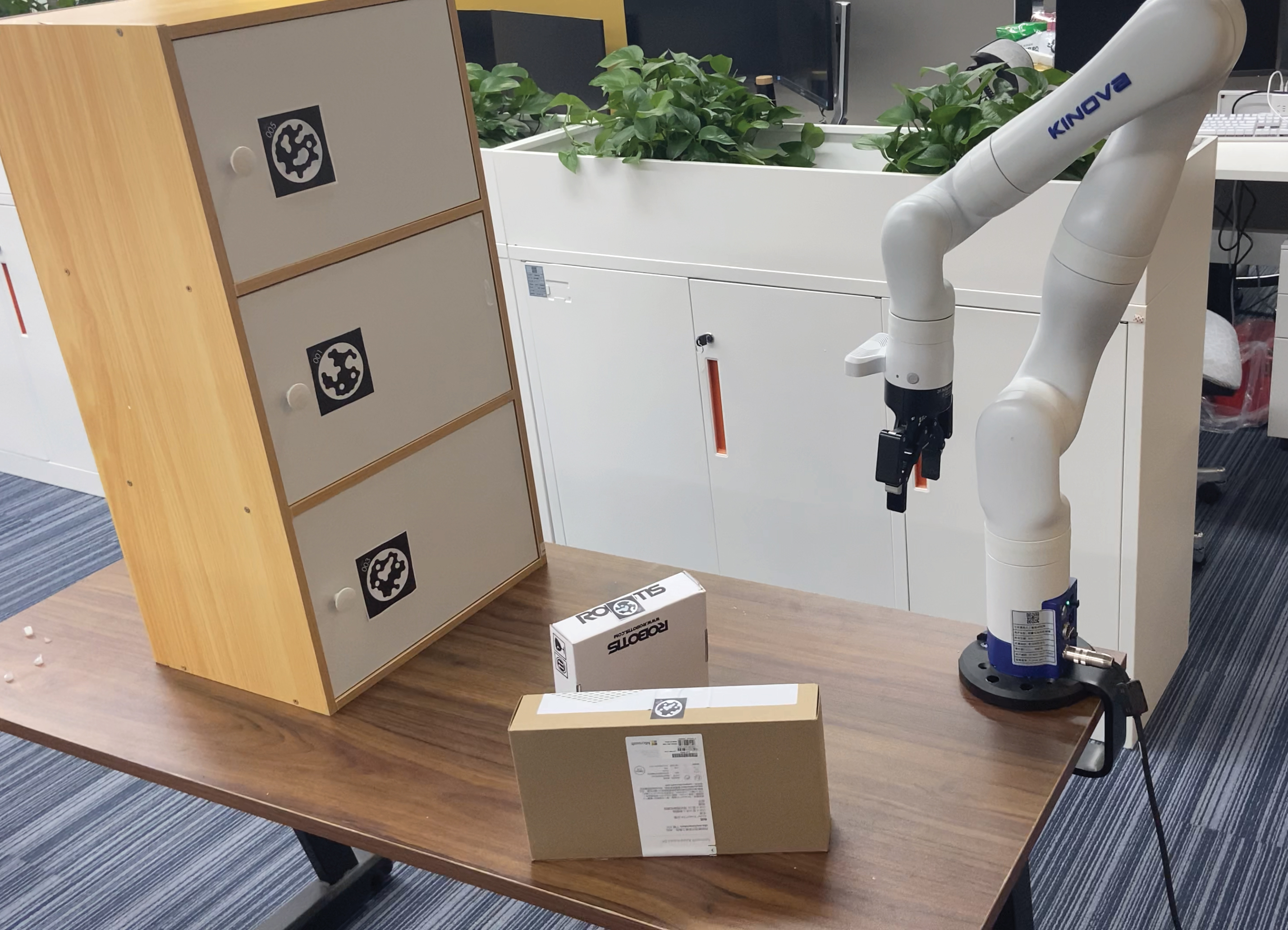}
        \caption{Setup}
        \label{fig:exp3_setup}
    \end{subfigure}%
    \begin{subfigure}[b]{0.3\linewidth}
        \centering 
        \includegraphics[width=\linewidth]{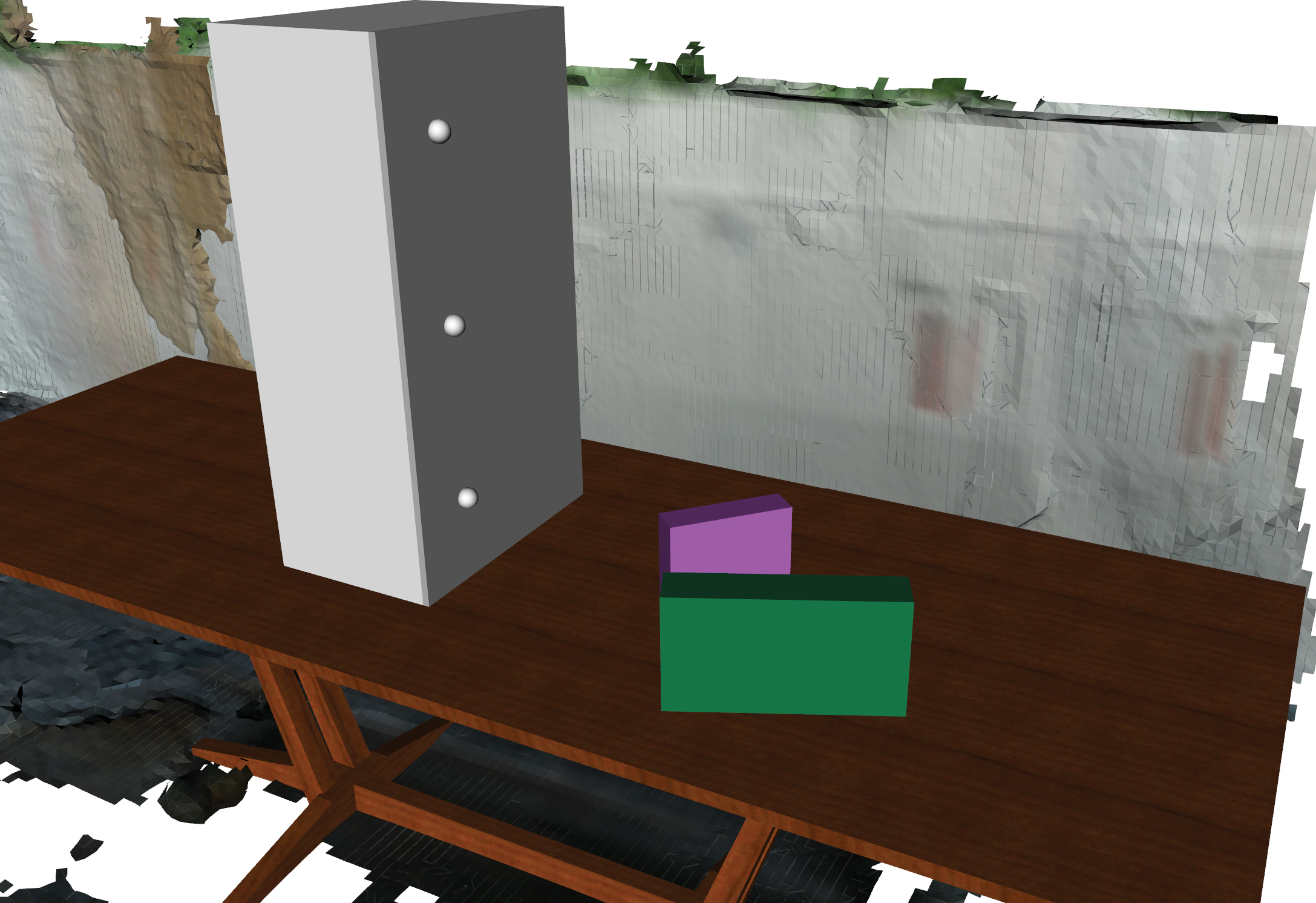}
        \caption{Reconstruct}
        \label{fig:exp3_reconstruct}
    \end{subfigure}%
    \begin{subfigure}[b]{0.3\linewidth}
        \centering
        \includegraphics[width=\linewidth]{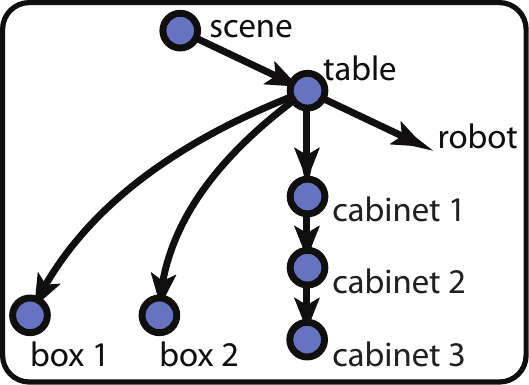}
        \caption{\ac{cg+}}
        \label{fig:exp3_cg}
    \end{subfigure}%
    \\%
    \begin{subfigure}[b]{\linewidth}
        \centering 
        \includegraphics[width=0.9\linewidth]{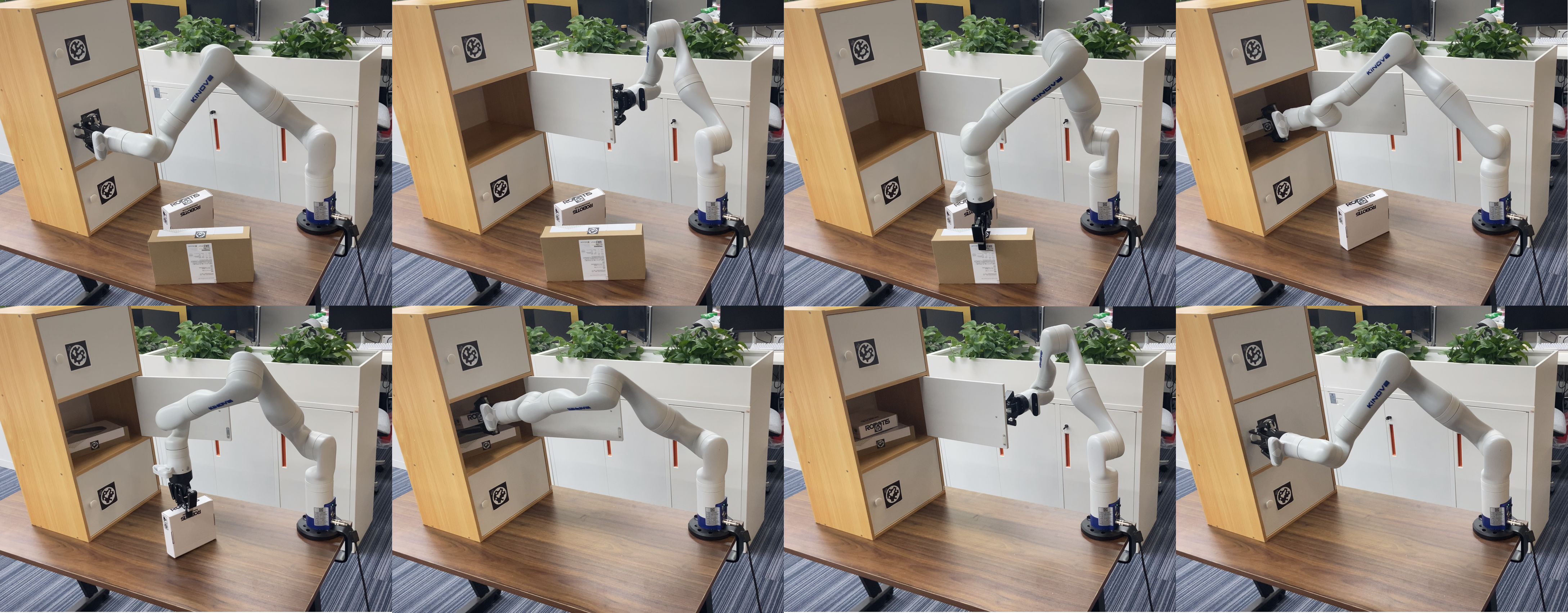}
        \caption{Planned action sequence.}
        \label{fig:exp3_plan}
    \end{subfigure}%
    \caption{Experiment of planning a complex sequential manipulation task on a contact graph generated from scene reconstruction.}
    \label{fig:exp3}
\end{figure}

We demonstrate \ac{cg+} in organizing information from scene reconstruction and performing manipulation planning in physical environment. The experiment is conducted with a Kinova Gen 3 manipulator on a table-top environment; see setup in \cref{fig:exp3_setup}, wherein the manipulator is tasked to place the two boxes on the table to the second cabinet. We reconstruct the scene~\cite{han2021reconstructing} and replace cabinets and boxes with CAD models (\cref{fig:exp3_reconstruct}). After generating contact graph (\cref{fig:exp3_cg}), the planned sequence on \ac{cg+} is shown in \cref{fig:exp3_plan}. As the two objects cannot fit into the cabinet side-by-side, the robot stacks the smaller object onto the larger one. Of note, the robot exhibits correct temporal order of executions for both cabinet opening and object stacking.

\section{Discussion and Conclusion}\label{sec:conclusion}

We tackled two challenges in manipulation planning for complex sequential tasks on contact graph\textemdash{}synthesizing plausible goal configurations and enforcing temporally correct graph edit operations. The former was addressed by a genetic algorithm that synthesizes the supporting relations among objects and a stochastic optimization method that produces objects' poses. The latter was converted to a topological sorting problem on the set of computed graph edit operations and imposed action temporal dependency defined by predicate-like attributes on \ac{cg+}. Our simulations and experiments demonstrate that the proposed planning scheme can scale up more efficiently compared with PDDLStream and handle spatially and temporally complex tasks whose planning domain could be hardly defined.

The proposed \ac{cg+} representation and the planning scheme are by no means a perfect solution for alleviating manual efforts completely in general settings. Rather, we aim at justifying the potential of developing planning scheme based on scene graph, which is fruitful for advancing robot autonomy by sharing a representation with perception and by relieving efforts in domain specification (\eg, those in \acp{pddl}). A future direction of the proposed framework is to handle uncertainty in robot perception and execution, which by itself is a large topic~\cite{garrett2020online,papallas2020online,ha2020probabilistic}. Scene graph representations, however, may afford new perspectives toward this problem. 

% \clearpage
\setstretch{0.91}
\fontsize{5}{5}\selectfont
\bibliographystyle{ieeetr}
\bibliography{reference}

\end{document}